\pdfoutput=1

\documentclass[11pt]{article}
\usepackage{amsmath}
\usepackage{inconsolata}

\usepackage[preprint]{acl}
\usepackage{longtable}
\usepackage{times}
\usepackage{latexsym}
\usepackage{multirow}
\usepackage{tcolorbox}
\usepackage{tabularx}

\usepackage{booktabs} 
\usepackage[T1]{fontenc}

\usepackage[utf8]{inputenc}
\usepackage{hyperref}

\usepackage{multirow}        
\PassOptionsToPackage{table}{xcolor}
\usepackage{xcolor}      

\usepackage{wrapfig}
\usepackage{floatflt}
\usepackage{color, colortbl}

\usepackage[inline,shortlabels]{enumitem}

\usepackage{lineno}

\definecolor{darkblue}{rgb}{0, 0, 0.5}
\hypersetup{colorlinks=true, citecolor=darkblue, linkcolor=darkblue, urlcolor=darkblue}
\definecolor{Highlight}{rgb}{0.92,0.94,1} 

\usepackage{microtype}

\usepackage{inconsolata}

\usepackage{graphicx}

\usepackage{todonotes}
\usepackage{cleveref}
\crefname{section}{\S}{\S}
\crefname{table}{Table}{Tables}
\crefname{figure}{Fig.}{Figs.}
\crefname{algorithm}{Alg.}{}
\crefname{ALC@unique}{Line}{Lines}
\crefname{equation}{Eq.}{Eqs.}
\crefname{appendix}{App.}{Apps.}
\crefformat{section}{\S#2#1#3} 

\usepackage{todonotes}

\title{DRP: \underline{D}istilled \underline{R}easoning \underline{P}runing with Skill-aware Step Decomposition for Efficient Large Reasoning Models}

\author{
Yuxuan Jiang$^{1}$ \quad Dawei Li$^{2}$ \quad Francis Ferraro$^{1}$ \\
$^{1}$University of Maryland, Baltimore County \\
$^{2}$Arizona State University \\
\texttt{yuxuanj1@umbc.edu}
}

\begin{document}
\maketitle

\begin{abstract}


While Large Reasoning Models (LRMs) excel at complex tasks via long Chain-of-Thought (CoT) reasoning, their outputs are often excessively verbose, leading to inefficiency. This problem is amplified when the student’s long-form reasoning mismatches the concise outputs of smaller teacher models—common in LLM distillation to avoid using costly large teachers. To address this issue, we propose \textbf{Distilled Reasoning Pruning (DRP)}, a hybrid framework that combines inference-time pruning with tuning-based distillation. DRP leverages a teacher model to perform \textit{skill-aware} step decomposition and pruning, then distills the refined reasoning paths into a student model, enabling efficient and accurate reasoning. Across challenging math datasets, DRP significantly reduces token usage without sacrificing accuracy—for instance, cutting tokens on GSM8K from 917 to 328 while improving accuracy from 91.7\% to 94.1\%, and reducing AIME tokens by 43\% with no performance drop. Further analysis shows that aligning training CoT structure with the student’s capacity is key to effective knowledge transfer. Code is available at: \href{https://github.com/YuxuanJiang1/DRP}{https://github.com/YuxuanJiang1/DRP}

\end{abstract}

\section{Introduction}






Although Large Reasoning Models (LRMs)~\cite{xu2025towards}, like OpenAI's o1~\cite{o1} and DeepSeek-R1~\cite{guo2025deepseek}, have advanced the state of the art in complex reasoning tasks~\cite{li2025system}, 
a critical limitation of these models is their tendency toward \textit{overthinking}—the generation of excessively verbose reasoning trajectories containing redundant or unnecessary steps~\cite{chen2024not,cui2025stepwise,fu2024efficiently}. This can lead to substantial inference overhead and misguide the model toward incorrect conclusions~\cite{sui2025stop}. %

\begin{figure}[t]
    \centering
    \includegraphics[width=1\linewidth]{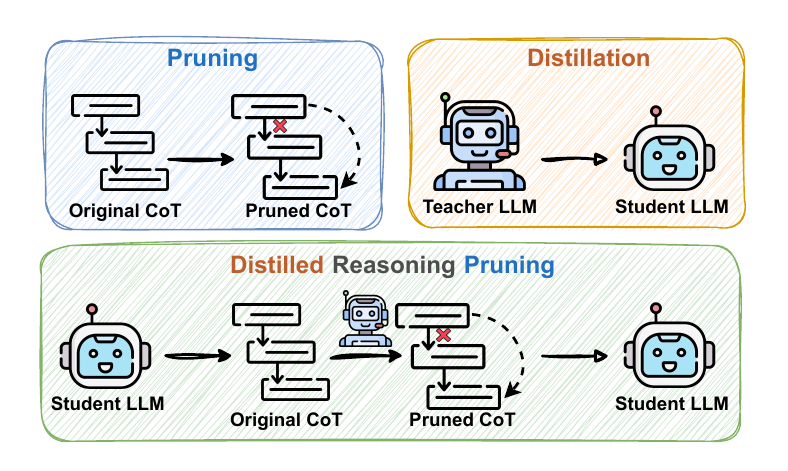}
    \caption{An overview of our proposed Distilled Reasoning Pruning (DRP) framework, which unifies pruning and distillation. Unlike traditional distillation, DRP uses a teacher LLM to prune the student model’s Long-CoT reasoning chains into concise CoTs, which are then distilled back into the student. This design addresses the reasoning style mismatch between verbose student models and concise teacher models, improving efficiency without sacrificing accuracy.}
    \label{revise}
\end{figure}

Existing solutions predominantly follow two paradigms: \textit{inference-time pruning}, which attempts to terminate generation early to avoid redundant reasoning steps~\cite{cui2025stepwise,muennighoff2025s1, fu2024efficiently, zeng2025revisiting}, and \textit{distillation-based compression}, where smaller models are trained on teacher-generated reasoning paths to imitate the concise reasoning behavior of larger models~\cite{tan2024large,zhu2024improving,xu2025twt}. However, both approaches can be at the cost of accuracy: pruning methods risk prematurely halting the reasoning process, while distillation methods tend to underperform due to the learnability gap~\cite{xu2024stronger,li2025small}. This gap becomes especially pronounced when the teacher adopts a Short-CoT style (concise, polished reasoning), and the student follows a Long-CoT style (verbose reasoning with reflective self-corrections). Such style mismatch introduces a compatibility issue that hinders effective learning, as discussed in prior work~\cite{xu2024stronger}.


\begin{figure*}
    \centering
    \includegraphics[width=1\linewidth]{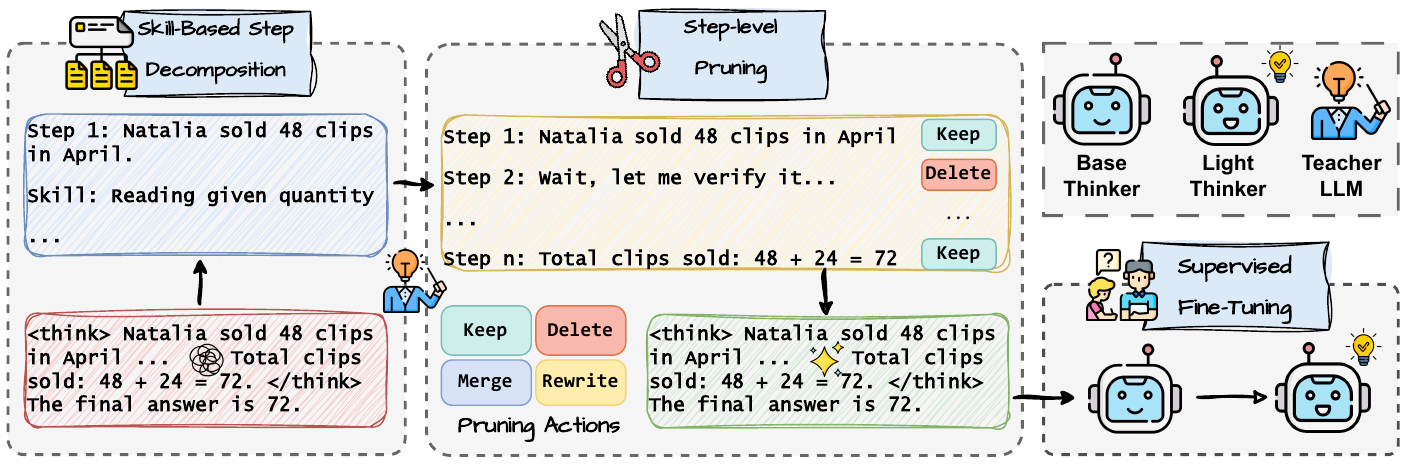}
    \caption{Overview of DRP framework. The student model generates Long-CoT reasoning traces, which are first decomposed into skill-based steps, then pruned and refined with help from a teacher LLM (e.g., GPT-4o). The concise CoTs align better with the student’s learning, improving efficiency without sacrificing accuracy.
}
    \label{fig:enter-label}
\end{figure*}

To address these limitations, we propose \textbf{Distilled Reasoning Pruning (DRP)} (\cref{revise}), a hybrid framework that combines the strengths of both pruning and distillation (student/teacher) paradigms. %
In particular, rather than simply distilling down a response from the teacher model, we use the teacher model to prune an initial, lengthy, CoT reasoning trajectory from the student model. To facilitate this, and to encourage shorter yet informative resulting pruned trajectories, we introduce a \textbf{skill-based step decomposition} method. The teacher model uses this to prune the trajectory, which produces more stable and semantically coherent reasoning units. %
Unlike prior methods that rely solely on either teacher-generated~\cite{xu2025twt,zhu2024improving} or self-sampled~\cite{chen2024not,ma2025cot} concise trajectories for distillation, DRP takes advantage of the teacher’s pruning within the student model’s original reasoning structure. 
This design reduces the learnability gap and enables student models to achieve efficient reasoning without compromising performance.

Consider the example in \cref{fig:enter-label}. For this word problem, the student model generates its initial CoT trajectory (shown in the \texttt{<think>} block). This trajectory is provided to a teacher model, which segments that CoT into steps, with a high-level description of the \textit{skill} that that step demonstrates or exercises. The teacher model then prunes and curates these skill-segmented steps, such as by merging similar ones (if the student was redundant or verbose) or deleting steps (e.g., for backtracking). This pruned CoT trajectory is then provided to the student model for supervised fine-tuning. %



We conduct extensive experiment in various student models, teacher models and mathematical datasets. DRP consistently improves the efficiency and accuracy of student models---even on significantly harder out-of-distribution tasks. Our ablation study demonstrates that DRP substantially outperforms direct distillation approaches. In addition, we show that our framework consistently benefits from various teacher models. We believe all these findings and insights from DRP can benefit future works in efficient reasoning in LRMs. Our key contributions are as follows:

\begin{itemize}
  \item We propose \textbf{Distilled Reasoning Pruning}, a novel framework that unifies step-level pruning and distillation to improve both reasoning efficiency and accuracy in small-scale LLMs.
  \item We introduce a \textbf{skill-based step decomposition} method that segments reasoning traces into semantically coherent and functionally aligned units, providing a stable foundation for pruning and supervision.
  \item We point out that effective training CoTs should be both informative and structurally consistent with the student's reasoning process, thereby facilitating knowledge transfer and bridging the capability gap between teacher and student models.
\end{itemize}

\section{Related Work}

\subsection{The Overthinking Problem}

Large reasoning models , such as OpenAI o1~\cite{o1} and DeepSeek-R1~\cite{guo2025deepseek}, are a subclass of LLMs trained to iteratively generate and refine intermediate steps~\cite{chen2025towards,sui2025stop,yu2025chain}, internalizing Chain-of-Thought (CoT) reasoning~\cite{tong2024can}. This leads to strong performance on complex tasks in math and code. However, LRMs—especially smaller ones—often overgenerate verbose reasoning chains with unnecessary tracebacks and redundant paths~\cite{chen2024not,cui2025stepwise,fu2025reasoning}, increasing token usage and causing reasoning drift that may harm accuracy~\cite{hou2025thinkprune,sui2025stop}. 

Our method addresses this \textit{overthinking} by pruning redundant steps and promoting concise, effective reasoning.

\subsection{Token-Efficient Reasoning Methods}

Current token-efficient reasoning methods fall into three categories~\cite{sui2025stop}:

\begin{enumerate*}[(1)]
\item \textbf{Prompt-based methods} constrain token budgets at the prompt level to encourage brevity without retraining. TALE~\cite{han2024token}, for example, estimates per-instance budgets to reduce output length. However, these methods depend on handcrafted prompts and struggle with complex tasks.

\item \textbf{Supervised fine-tuning (SFT)} trains models on compressed traces to internalize efficiency~\cite{xia2025tokenskip, munkhbat2025self}, as in CoT-Valve~\cite{ma2025cot}. These methods typically require task-specific data and retraining.

\item \textbf{Reinforcement learning (RL)} introduces rewards to penalize long outputs~\cite{team2025kimi, chen2024not}, sometimes with early-exit mechanisms~\cite{muennighoff2025s1, fu2024efficiently, zeng2025revisiting,dai2025s,yang2025dynamic}. ThinkPrune~\cite{hou2025thinkprune} sets target lengths and tightens constraints over time.
\end{enumerate*}

Despite these advances, many approaches reduce accuracy—especially on out-of-domain (OOD) tasks—when optimizing for brevity. By contrast, our framework uses external teacher models to perform \textit{skill-aware pruning}, reducing tokens while improving robustness under distribution shift.

\subsection{LLM Self-Refinement}

Recent work explores \textit{self-refinement}, where a model iteratively revises its own outputs to improve accuracy~\cite{madaan2023self,li2024selective}, often through self-feedback or selective re-generation.

In contrast, our approach introduces external teacher supervision to prune redundant steps during reasoning. Rather than focusing on iterative correction of final answers, our \textit{skill-aware pruning} directly optimizes the reasoning structure, thereby improving both token efficiency and reasoning quality.

\section{Methodology}
We propose \textbf{D}istilled \textbf{R}easoning \textbf{P}runing (DRP), a method to improve the efficiency of Long-CoT student models by refining their reasoning traces using a concise Short-CoT teacher. This asymmetric setup is key: the student (e.g., R1-Distill-Qwen-7B) generates verbose, reflective reasoning, while the teacher (e.g., GPT-4o) produces polished, compact CoTs. To bridge this gap, we let the student generate initial traces, which are selectively revised under teacher guidance.

As shown in Figure~\ref{fig:enter-label}, DRP involves three stages: 
\begin{enumerate*}[(1)]
\item decomposing reasoning into fine-grained, skill-based steps;
\item pruning and rewriting via a teacher LLM; and 
\item fine-tuning the student on the revised traces.
\end{enumerate*}
This process yields token-efficient supervision that enhances both accuracy and reasoning efficiency.

\subsection{Skill-Based Step Decomposition}

To solve a math problem, a reasoning model produces a response consisting of two parts: a structured reasoning trace \(T\) enclosed in \texttt{<think>} \texttt{</think>} tags, and a final answer summarization \(A\). We denote the response as \(R=(T,A)\). For example, in \cref{fig:enter-label}, the trace may include ``Natalia sold 48 clips in \dots Total clips sold: 48 + 24 = 72,'' while the answer summarization is ``The answer is 72.''

We extract the trace \(T\) and prompt a teacher model to decompose it into non-overlapping segments, each aligned with a functional reasoning skill (e.g., arithmetic, comparison, logical inference). This skill-based segmentation supports step-level pruning and distillation. Compared with naïve sentence splitting, it yields more stable step boundaries, preserves semantic coherence, and provides finer-grained supervision signals (see \cref{rq1}). The segmentation quality is validated through pairwise evaluation with Gemini 2.0 Flash; results and examples appear in Appendix~\ref{skill eval}.

Formally, given a response \(R=(T,A)\), we define a decomposition function \(D\):
\[
D(T) \mapsto \{(s_1, k_1), (s_2, k_2), \ldots, (s_m, k_m)\},
\]
where each \(s_i\) is a contiguous token span representing one reasoning step, and \(k_i\) is the corresponding skill label assigned by the teacher model. For instance, ``Natalia sold 48 clips in April'' corresponds to the skill ``Reading given quantity.'' Skills also cover broader tasks such as ``Algebraic representation'' or ``Interpreting fractions of a subset.'' Complete decomposition prompts are provided in Appendix~\ref{appendix:prompts}.








\subsection{Step-Level Pruning}
\label{action}

For each reasoning step–skill pair \( (s_i, k_i) \), the teacher model is prompted to revise the step without altering the essential structure or logical intent of the original reasoning. The teacher selects one of the following actions for each step:
\begin{itemize}[leftmargin=.75em,itemsep=0em]
    \item \textbf{Keep:} Retain the step unchanged.
    \item \textbf{Delete:} Remove the step if it is redundant or uninformative.
    \item \textbf{Rewrite:} Replace the step with a more concise version conveying the same logic.
    \item \textbf{Merge:} Combine the step with adjacent ones if they form a coherent atomic unit.
\end{itemize}
\noindent This yields a revised step \( \hat{s}_i = \text{Revise}(s_i) \), and a pruned reasoning trace:
$\hat{T} = \{\hat{s}_1, \hat{s}_2, \dots, \hat{s}_{m'}\}$, where $m' \leq m$, which reduces redundancy and increases the overall information density.

Finally, the teacher model rewrites \( \hat{T} \) into a fluent, coherent reasoning trace that preserves the tone and speaking style of the student model.  
To ensure consistency between the revised reasoning and the final answer, we prompt the teacher to optionally revise the original answer segment \( A \), yielding an updated final answer summarization \( \hat{A} \). The final output becomes a concatenation \( \hat{R} = (\hat{T}, \hat{A}) \), which we use as the target for supervised fine-tuning. Complete prompt are in Appendix~\ref{appendix:prompts}.

\begin{tcolorbox}[colback=gray!5, colframe=gray!50, title=Self-Revision Prompt]
{\slshape 
Given a list of reasoning steps labeled with their respective skills, your task is to evaluate and revise each step according to one of the four ACTIONs.

Ensure the resulting reasoning path is concise, fluent, and logically sound. At the end, synthesize the revised steps into a coherent explanation that matches the speaker’s tone.
}
\end{tcolorbox}





\subsection{Supervised Fine-Tuning}

We construct the training dataset using pairs \( (x, \hat{R}) \), where \( x \) is the input question and \( \hat{R} \) is the complete revised response. 

We fine-tune the model using teacher-forced decoding to encourage the generation of concise reasoning traces. The training objective minimizes the negative log-likelihood of the revised response:
\[
\mathcal{L}_{\text{SFT}} = -\sum_{i=1}^{n} \log P_{\theta}(y_i \mid x, y_{<i}),
\]
where \( \{y_1, \dots, y_n\} \) are the tokens in \( \hat{R} \), and \( \theta \) denotes the model parameters. This supervision enables the model to internalize skill-aligned, token-efficient reasoning strategies while maintaining consistency with the final answer.

\begin{table*}[htbp]
\centering
\small
\resizebox{1\textwidth}{!}{%
\begin{tabular}{l|cc|cc|cc|cc}
\toprule
\textbf{Method} 
& \multicolumn{2}{c|}{\textbf{GSM8K}} 
& \multicolumn{2}{c|}{\textbf{MATH500 (OOD)}} 
& \multicolumn{2}{c|}{\textbf{AIME24 (OOD)}} 
& \multicolumn{2}{c}{\textbf{AMC (OOD)}} \\
& Pass@1 & \#Tokens 
& Pass@1 & \#Tokens 
& Pass@1 & \#Tokens 
& Pass@1 & \#Tokens \\
\midrule

\multicolumn{9}{l}{\textbf{R1-Distill-Qwen-1.5B}} \\
\midrule
Base & 70.7\% & 1443 & 80.4\% & 3276 & 6/30 & 10484 & 23/40 & 6516 \\
+TALE & 70.1\% & 1170 & 76.2\% & 3107 & 6/30 & 8915 & 22/40 & 6235 \\
+Cot Valve & 70.4\% & 805 & 76.5\% & 2705 & 7/30 & \textbf{5601} & 22/40 & 4574 \\
+Thinkprune & 80.0\% & \textbf{712} & 79.2\% & \textbf{2006} & 9/30 & 5745 & 25/40 & \textbf{3291} \\
\rowcolor{Highlight}
+DRP & \textbf{83.4\%} & 721 {\scriptsize\textcolor{olive}{(-50\%)}} & \textbf{82.0\%} & 2122 {\scriptsize\textcolor{olive}{(-35\%)}} & \textbf{10/30} & 6135 {\scriptsize\textcolor{olive}{(-42\%)}} & \textbf{27/40} & 3657 {\scriptsize\textcolor{olive}{(-44\%)}} \\

\midrule
\multicolumn{9}{l}{\textbf{R1-Distill-Qwen-7B}} \\
\midrule
Base & 91.7\% & 917 & 92.4\% & 2486 & 15/30 & 8674 & 31/40 & 4845 \\
+TALE & 91.0\% & 522 & 91.6\% & 2530 & 10/30 & 8602 & 31/40 & 3998 \\
+Cot Valve & 90.8\% & 364 & 89.4\% & 1975 & 13/30 & 6315 & 30/40 & \textbf{3157} \\
\rowcolor{Highlight}
+DRP & \textbf{94.1\%} & \textbf{328} {\scriptsize\textcolor{olive}{(-64\%)}} & \textbf{93.0\%} & \textbf{1781} {\scriptsize\textcolor{olive}{(-28\%)}} & \textbf{15/30} & \textbf{4966} {\scriptsize\textcolor{olive}{(-43\%)}} & \textbf{33/40} & 3258 {\scriptsize\textcolor{olive}{(-33\%)}} \\
\bottomrule

\end{tabular}
} 
\caption{Pass@1 accuracy and average token usage on R1-Distill-Qwen models across various math benchmarks, comparing our DRP method with Cot Valve~\cite{ma2025cot}, TALE~\cite{han2024token}, and ThinkPrune~\cite{hou2025thinkprune}.}
\label{main_results}
\end{table*}

\section{Experimental Setup}

\subsection{Datasets}
Our training corpora consist of the training split of GSM8K~\cite{gsm8k} and the full PRM12K~\cite{lightman2023let} dataset. From these, we generate initial reasoning paths, which are then processed through skill-based step decomposition and teacher-guided step-level pruning to create supervision signals for fine-tuning.
To evaluate complex mathematical reasoning and generalization ability, we select a broad set of out-of-domain benchmarks including MATH500~\cite{math}, AIME24~\cite{aime2024}, and AMC23~\cite{amc2024}.

\subsection{Models}

We use DeepSeek-R1-Distill-Qwen-7B and DeepSeek-R1-Distill-Qwen-1.5B~\cite{guo2025deepseek} as student models for supervised fine-tuning. Both are distilled variants of DeepSeek-R1 optimized for efficient inference. Our primary teacher model is GPT-4o~\cite{openai2024gpt4o}, which performs step-level decomposition and pruning to generate supervision signals. For ablation studies, we additionally explore alternative teacher models, including Gemini 2.0 Flash~\cite{google2025gemini}, Deepseek V3~\cite{liu2024deepseek} and ChatGPT~\cite{openai2022chatgpt}.

\subsection{Compared Methods}
\label{sec:baselines}
We select three representative methods that span the major paradigms for token-efficient reasoning:
\begin{enumerate*}[(1)]
\item \textbf{TALE}~\cite{han2024token}: a prompt-based method incorporates a soft token budget constraint into the prompt to encourage concise generation. %
\item \textbf{CoT-Valve}~\cite{ma2025cot}: a SFT-based method which enerates multiple chains-of-thought of varying lengths for the same problem, and performs supervised fine-tuning in multiple rounds—each time using shorter CoTs as training targets. %
\item \textbf{ThinkPrune}~\cite{hou2025thinkprune}: a tuning-based method uses reinforcement learning to iteratively reduce chain-of-thought length by optimizing under a target token constraint.
\end{enumerate*}

\subsection{Implementation Details}

Supervised fine-tuning is performed using the \textsc{LLaMA-Factory}\footnote{\url{https://github.com/hiyouga/LLaMA-Factory}} framework with LoRA adaptation. All models are trained for 3 epochs with cosine learning rate scheduling. Full hyperparameters are listed in Appendix~\ref{app:sft}.

\subsection{Evaluation Protocol}

We use the \textsc{lm-evaluate-harness}\footnote{\url{https://github.com/EleutherAI/lm-evaluation-harness}} framework for unified evaluation across tasks. Each model is evaluated in a zero-shot setting. For inference, we use the vLLM backend with the maximum generation length set to 131{,}072 tokens—the upper limit supported by the Qwen models.

\subsection{Evaluation Metrics}
\label{metrics}

We report \textbf{Pass@1} as the accuracy metric, averaged over five independent runs to account for randomness in decoding. For efficiency, we measure the number of reasoning tokens generated per completion using the HuggingFace-compatible Qwen tokenizer.\footnote{\url{https://huggingface.co/Qwen/Qwen-tokenizer}}

We also observe that models occasionally fall into degenerate loops, repeatedly generating parts of their responses until reaching the maximum generation limit (e.g., 130k tokens), far exceeding the typical average length (e.g., 5k). In most of these cases, the model fails to answer correctly. Such outliers significantly inflate the average token count.

To mitigate their impact, we set a cutoff threshold of 12k tokens, which empirically covers 99\% of correct responses across all benchmarks for the models we evaluate. Detailed token length distributions by task are provided in Appendix~\ref{outlie}.

\section{Main Results}

\subsection{Accuracy and Token Efficiency}
\label{sec:main_result}
Table~\ref{main_results} presents the main results on our DRP and other compared methods. Our key findings are:

\paragraph{DRP consistently reduces token usage across all benchmarks and model sizes.}  
Our proposed DRP method achieves substantial reductions in average token usage on both in-domain and out-of-domain tasks. On GSM8K, DRP reduces token count by up to 64\% with the 7B model. For out-of-domain datasets, DRP yields 28\%–44\% reductions across all benchmarks, demonstrating strong generalization beyond the training distribution.

\paragraph{DRP improve the accuracy by mitigating the over-thinking problem in LRMs.}  
Despite significantly reducing token usage, DRP preserves or improves Pass@1 accuracy on nearly all benchmarks. Notably, DRP improves accuracy even on harder datasets such as AMC and MATH500. The only exception is AIME24 under the 7B setting, where accuracy remains unchanged, suggesting the inherent difficulty of this benchmark.

\paragraph{Accuracy gains are more pronounced for smaller models.}  
DRP delivers particularly strong improvements with the 1.5B model. On GSM8K, accuracy increases by 12.7\%, while on AIME24 and AMC, DRP answers 4 more problems correctly compared to the base model. This indicates DRP’s effectiveness in compensating for limited model capacity through more efficient supervision.

\subsection{Comparison to Prior Work}

We compare our approach with three representative baselines covering prompt-based, SFT-based, and RL-based token-efficient reasoning strategies (details can be found in Section~\ref{sec:baselines}).

\paragraph{Prompt-based methods offer limited control on Complex Reasoning Tasks.}  
TALE is simple to implement and demonstrates moderate effectiveness on short-answer tasks such as GSM8K. It achieves token reductions with minimal accuracy degradation (less than 1\%) on both model sizes. However, its effectiveness diminishes significantly on more challenging benchmarks. For example, on AIME24 with the 7B model, TALE causes a notable performance drop.  Overall, TALE's prompt constraints provide limited control over fine-grained reasoning behaviors, which we hypothesize leads to its suboptimal performance on tasks requiring deeper or longer reasoning chains. In contrast, our DRP method remains effective even on such difficult tasks.

\paragraph{Existing SFT Methods struggle to balance accuracy and efficiency.}  
CoT-Valve achieves consistent token reductions across benchmarks by training on compressed CoT. However, this often comes with accuracy trade-offs. For instance, on MATH500 and AMC, we observe accuracy degradation despite reduced token usage. By comparison, our DRP method achieves both stronger compression and accuracy improvements across most benchmarks. Notably, on AMC (1.5B), DRP improves accuracy from 22/40 to 27/40 while also reducing tokens from 4574 to 3657.

\paragraph{Fine-grained supervised pruning improves reasoning accuracy more effectively than RL-based methods.}
ThinkPrune demonstrates strong performance on the 1.5B model, suggesting that pruning is effective in eliminating distracting or redundant reasoning steps, particularly for smaller-capacity models. Notably, it achieves solid gains on benchmarks like AIME24 and AMC. However, our DRP method yields higher accuracy improvements, likely due to its use of high-quality teacher-guided pruning. While ThinkPrune achieves slightly better compression on a few tasks—due to its explicit optimization for token length, DRP achieves notably higher accuracy across the board, while still maintaining competitive compression rates.

\subsection{Token Usage Distribution Shift}
\begin{figure}[t]
    \centering
    \includegraphics[width=7.5cm]{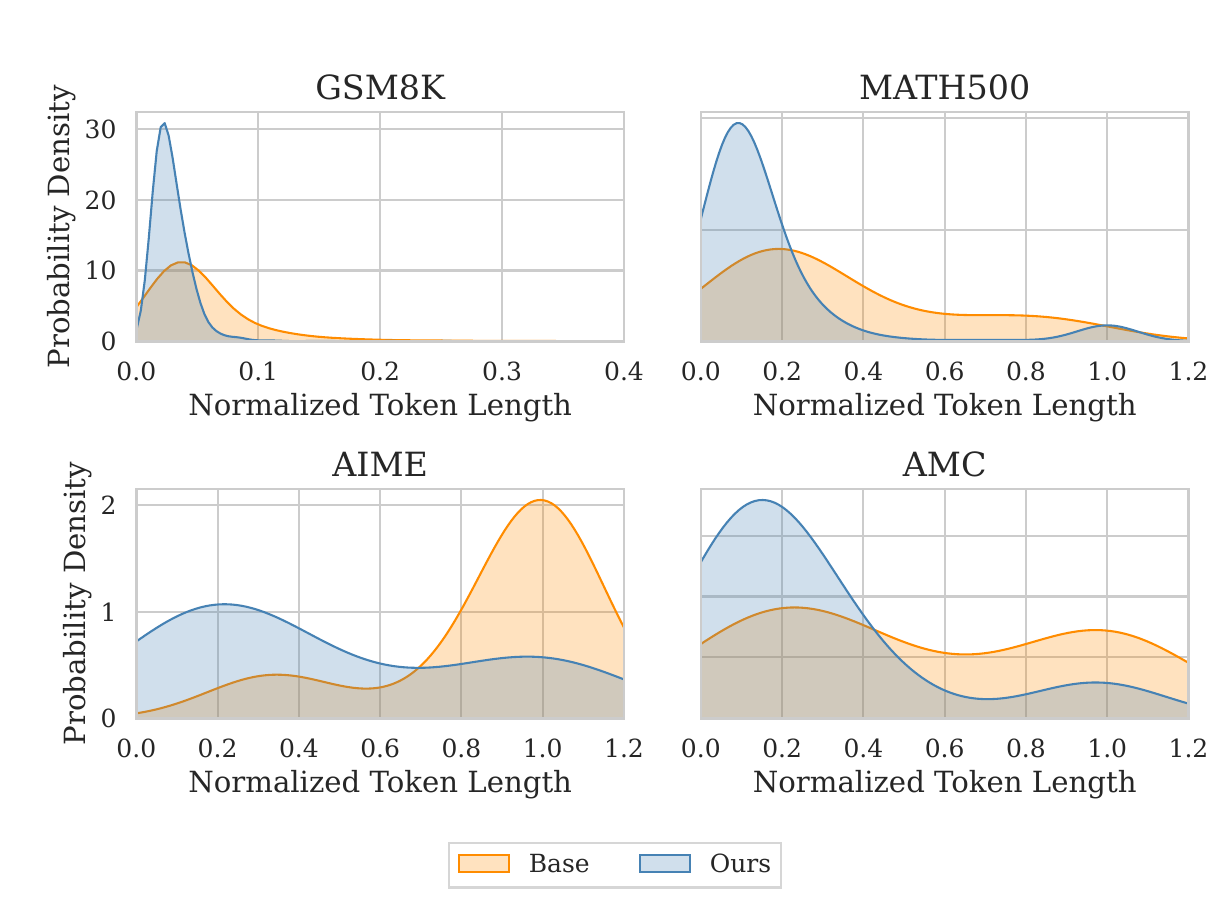}
    \caption{Normalized token length distributions across GSM8K, MATH500, AIME24, and AMC before and after SFT using the DeepSeek-R1-Distill-Qwen-7B model. The horizontal axis indicates the normalized token length (token count divided by the maximum allowed length), and the vertical axis represents the probability density. Blue curves correspond to our method (DRP), and orange curves denote the baseline. The reduction in long-tail completions and high-token outliers indicates that DRP mitigates verbose and degenerate reasoning failures, resulting in more robust and efficient inference.}
\label{fig:token_shift_distribution}

\end{figure}

Figure~\ref{fig:token_shift_distribution} shows the distribution of normalized token lengths before and after applying DRP across all benchmarks using R1-Distill-Qwen-7B. Detailed observations are discussed below:

\paragraph{Concise Reasoning Across Tasks.}
DRP effectively compresses the overall reasoning length across all benchmarks. The main density of the token distribution shifts leftward, especially on harder out-of-domain tasks like MATH500, AIME24, and AMC. This demonstrates that DRP encourages more concise reasoning behavior.

\paragraph{Eliminating Verbose and Degenerate Reasoning Failures.}
DRP significantly reduces long-tail completions that typically result from verbose or degenerate reasoning. On AMC, the baseline exhibits a bimodal distribution, where the secondary peak reflects unnecessarily lengthy yet still valid reasoning paths. More critically, we observe a clear drop in density near the upper token limit—especially in AIME and AMC—indicating that DRP mitigates degenerate cases where the model previously generated excessively long or looping outputs due to failure to converge (Section~\ref{metrics}). This results in improved \textit{robustness}, defined here as the model’s ability to terminate reasonably when it cannot reach a correct solution. 

\paragraph{Further Compression of Already Efficient Reasoning Paths.}
On datasets such as GSM8K and MATH500, where baseline models already produce relatively short completions, DRP still yields measurable compression gains. This indicates that DRP not only removes verbosity but also optimizes reasoning even in high-performing regimes.

\begin{table}[t]
\centering
\small
\begin{tabular}{lccc}
\toprule
\textbf{Dataset} & \textbf{Method} & \textbf{Accuracy} & \textbf{Tokens} \\
\midrule
\multirow{4}{*}{GSM8K} 
& 7B Base        & 91.7\%  & 917  \\
& No decomposing & 91.0\%  & 434  \\
& Default        & 92.7\%  & 350  \\
& DRP            & 94.1\%  & 328  \\
\midrule
\multicolumn{4}{c}{\textit{OOD Tasks}} \\
\midrule
\multirow{4}{*}{MATH500} 
& 7B Base        & 92.4\%  & 2486 \\
& No decomposing & 88.6\%  & 2102 \\
& Default        & 92.0\%  & 1905 \\
& DRP            & 93.0\%  & 1781 \\
\midrule
\multirow{4}{*}{AIME24} 
& 7B Base        & 15/30   & 8674 \\
& No decomposing & 13/30   & 6201 \\
& Default        & 14/30   & 4678 \\
& DRP            & 15/30   & 4966 \\
\midrule
\multirow{4}{*}{AMC} 
& 7B Base        & 31/40   & 4845 \\
& No decomposing & 29/40   & 4028 \\
& Default        & 31/40   & 4975 \\
& DRP            & 33/40   & 3258 \\
\bottomrule
\end{tabular}
\caption{Comparison of 7B base model, no decomposition, default step segmentation, and our DRP method across in-distribution and out-of-distribution math benchmarks.}
\label{split_comp}
\end{table}

\section{Ablation Studies}
We conduct three ablation experiments to understand the contribution of each component in our DRP framework separately.

\subsection{Skill-based Decomposition vs. Default Step Split}
\label{rq1}

\textbf{RQ1: Does skill-based decomposition improve downstream learning compared to default step segmentation?}

To evaluate the impact of our skill-based decomposition, we compare three variants under the same DRP framework: (1) skill-based segmentation, (2) default step-wise splitting without skill labels, and (3) no decomposition at all.

\begin{tcolorbox}[colback=gray!5, colframe=gray!50, title=Comparison of Decomposition Prompts]
\textbf{Skill-Based Prompt:}  
Segment the explanation into a sequence of clear, non-overlapping steps, where each step corresponds to exactly one atomic mathematical skill. These skills may include, for example, \textit{addition}, \textit{subtraction}, \textit{applying a formula}, \textit{interpreting a quantity}, \textit{simplifying}, \textit{checking a condition}, and so on.

\vspace{1ex}
\textbf{Default Prompt:}  
Segment the explanation into clear, non-overlapping steps.
\end{tcolorbox}

As shown in Table~\ref{split_comp}, skill-based segmentation consistently outperforms the other variants, improving accuracy (e.g., +2 on AMC) and reducing token usage more effectively (e.g., 3258 vs.\ 4975). This gain is partly due to finer granularity—our method produces 12.6 steps per example on GSM8K, compared to 8.3 with default splitting (see Appendix~\ref{skill_default}).

\begin{table*}
\centering
\small
\resizebox{16.0cm}{!}{\begin{tabular}{lcccccccc}
\toprule
\textbf{Method} 
& \multicolumn{2}{c}{\textbf{GSM8K}} 
& \multicolumn{2}{c}{\textbf{MATH500}} 
& \multicolumn{2}{c}{\textbf{AIME24}} 
& \multicolumn{2}{c}{\textbf{AMC}} \\
\cmidrule(lr){2-3} \cmidrule(lr){4-5} \cmidrule(lr){6-7} \cmidrule(lr){8-9}
& Accuracy & Tokens 
& Accuracy & Tokens 
& Accuracy & Tokens 
& Accuracy & Tokens \\
\midrule
7B-base student model
& 91.7\% & 917 
& 92.4\% & 2486 
& 15/30 & 8674
& 31/40 & 4845 \\
\quad with GPT-4o teacher
& 94.1\% & 328 
& 93.0\% & 1781 
& 15/30 & 4966 
& 33/40 & 3258 \\
\quad with Gemini 2.0 Flash teacher
& 93.2\% & 419
& 91.8\% & 2245 
& 14/30 & 4860
& 33/40 & 2835 \\
\quad with ChatGPT teacher
& 91.2\% & 386
& 90.2\% & 2122 
& 14/30 & 5265 
& 32/40 & 3101 \\
\quad with DeepSeek-V3 teacher
& 92.7\% & 340
& 90.4\% & 1908 
& 14/30 & 4640 
& 32/40 & 3136 \\
\bottomrule
\end{tabular}}
\caption{Impact of different teacher models on DRP performance. We compare GPT-4o, Gemini 2.0 Flash, ChatGPT, and DeepSeek-V3 as pruning teachers, evaluating downstream accuracy and average token usage.}
\label{teacher_comp}
\end{table*}

In contrast, skipping decomposition entirely leads to significant performance degradation across all benchmarks, confirming that explicit step segmentation is essential for effective pruning.
\textit{Notably, this structural benefit generalizes:} skill-based segmentation remains robust even on out-of-distribution tasks, yielding more transferable supervision.

\subsection{Structured Pruning vs. Direct Distillation}
\label{rq2}
\textbf{RQ2: Does the performance gain come from shorter CoTs or the structured pruning process?}

\begin{table}[t]
\centering
\small
\begin{tabular}{lccc}
\toprule
\textbf{Dataset} & \textbf{Method} & \textbf{Accuracy} & \textbf{Tokens} \\
\midrule
\multirow{3}{*}{GSM8K} 
& 7B Base & 91.7\% & 917 \\
& Distill & 90.7\% & 425 \\
& DRP     & 94.1\% & 328 \\
\midrule
\multicolumn{4}{c}{\textit{OOD Tasks}} \\
\midrule
\multirow{3}{*}{MATH500} 
& 7B Base & 92.4\% & 2486 \\
& Distill & 88.6\% & 2152  \\
& DRP     & 93.0\% & 1781 \\
\midrule
\multirow{3}{*}{AIME24} 
& 7B Base & 15/30 & 8674 \\
& Distill & 13/30 & 6417 \\
& DRP     & 15/30 & 4966 \\
\midrule
\multirow{3}{*}{AMC} 
& 7B Base & 31/40 & 4845 \\
& Distill & 28/40 & 4279 \\
& DRP     & 33/40 & 3258 \\
\bottomrule
\end{tabular}
\caption{Comparison between 7B base model, direct distillation from GPT-4o, and our DRP method across both in-distribution and OOD math benchmarks.}
\label{distill_comp}
\end{table}

To answer this, we compare DRP with a direct distillation baseline, where the student model is trained on concise CoT generated directly by GPT-4o. These responses are notably short---averaging \textbf{186 tokens}---compared to DRP's CoT average of \textbf{330 tokens}. Table~\ref{distill_comp} presents the results.
We observe that:
\begin{itemize}[itemsep=0.2em, topsep=0em, leftmargin=1.5em]
\item On the in-distribution GSM8K, direct distillation cuts tokens by over 50\% (425 vs.\ 917) with only a minor accuracy drop (90.7\% vs.\ 91.7\%), showing effectiveness on simpler tasks.
\item On OOD tasks like MATH500, AIME24, and AMC, accuracy drops sharply despite modest token savings (e.g., 88.6\% vs.\ 93.0\% on MATH500), indicating poor generalization.
\end{itemize}

These results show that shorter CoTs alone do not generalize well. Direct distillation often omits critical steps, harming transfer. In contrast, DRP removes redundancy while preserving a step-wise, skill-aligned structure, enabling more robust reasoning. This suggests a key insight: preserving structural depth—even with moderate length—outperforms naive compression. See Appendix~\ref{long_cot} for examples.

\subsection{Influence of Teacher Model Choice}
\label{rq3}
\textbf{RQ3: How does the choice of teacher model influence pruning effectiveness?}

To evaluate the sensitivity of our pruning framework to the choice of teacher model, we experiment with four different large language models (LLMs): GPT-4o, Gemini 2.0 Flash, ChatGPT, and DeepSeek-V3. As shown in Table~\ref{teacher_comp}, all teacher models yield consistent improvements in both accuracy and token efficiency over the base model, confirming the robustness and generalizability of our approach.

Among them, GPT-4o achieves the strongest performance, especially in token compression. Using DeepSeek-V3 as the teacher also leads to clear gains over the student baseline, with results comparable to other teacher models. This suggests that teacher–student family alignment is not critical for the effectiveness of our DRP method.

Overall, while stronger teachers can yield slightly more optimal pruning decisions, the differences remain modest. Our method generalizes well across teacher models and does not overfit to a specific model’s reasoning style. Notably, we observe greater variation in token usage than in accuracy, indicating that our skill-aware pruning retains its core benefits regardless of teacher strength.

\section{Conclusion}
We propose skill-based Distilled Reasoning Pruning, a framework that leverages pruned reasoning traces to distill smaller reasoning models. It outperforms direct distillation by effectively bridging the learnability gap between student and teacher models. The success of our skill-based step decomposition underscores the importance of fine-grained, consistent step segmentation as a strong foundation for pruning.

\section{Limitations}
While our method demonstrates strong performance on existing small-scale reasoning models, it remains uncertain how well it generalizes to other architectures. Currently, there is a limited number of publicly available small-sized LRMs with strong reasoning capabilities, making broad validation challenging. In addition, the training paradigms for reasoning models are evolving rapidly, and it is unclear whether the overthinking and inefficiency issues we target will persist in future model generations. We view our work as a step toward addressing current bottlenecks, but acknowledge that its relevance may shift as the landscape of LLM training continues to change.

\section{Ethics}
Our study uses the OpenAI and Google Gemini API on experiments above, and at no point do we access, or attempt to access, the true training data behind these models, or any underlying components of the systems.
\textbf{Risks}
The several datasets in our experiment are sourced from publicly available sources. However, we cannot guarantee that they are devoid of socially harmful or toxic language. We use ChatGPT~\footnote{\url{https://chatgpt.com/}} to correct grammatical errors in this paper.

\bibliography{custom}
\bibstyle{acl_natbib}

\appendix
\section{Outliers and Cutting Off}
\label{outlie}
\begin{figure}[h]
    \centering
    \includegraphics[width=7.5cm]{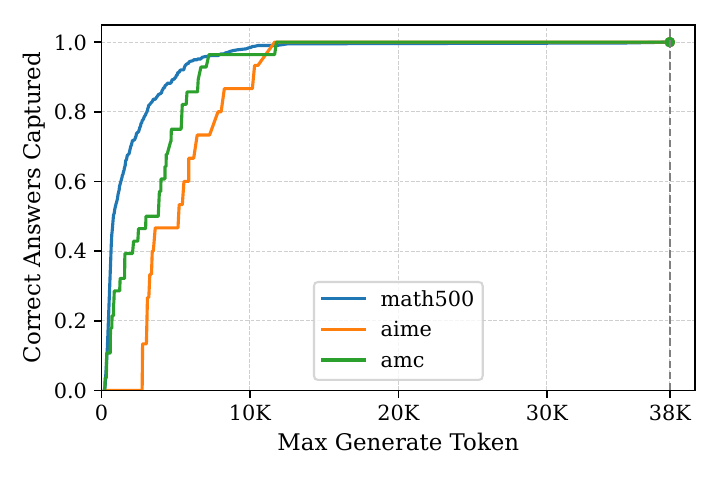}
    \caption{The x-axis denotes the model’s maximum generation length, and the y-axis shows the proportion of correct answers recovered within that budget, with the R1-Distill-Qwen-7B.}

    \label{capture}
\end{figure}

We can observe from Figure~\ref{capture} that most answers are captured by 11K tokens, though some require up to 38K, revealing a long-tail pattern.

\begin{table*}[htbp]
\centering

\resizebox{16.0cm}{!}{
\begin{tabular}{lcccccccc}
\toprule
\textbf{Method} 
& \multicolumn{2}{c}{\textbf{GSM8K}} 
& \multicolumn{2}{c}{\textbf{MATH500}} 
& \multicolumn{2}{c}{\textbf{AIME24}} 
& \multicolumn{2}{c}{\textbf{AMC}} \\
\cmidrule(lr){2-3} \cmidrule(lr){4-5} \cmidrule(lr){6-7} \cmidrule(lr){8-9}
& Accuracy & Tokens 
& Accuracy & Tokens 
& Accuracy & Tokens 
& Accuracy & Tokens \\
\midrule
Baseline (no cutoff) 
& 91.7\% & 917
& 92.5\% & 5435
& 15/30 & 54339
& 32/40 & 18736 \\

Baseline (12k cutoff) 
& 91.7\% & 917 
& 92.4\% & 2486 
& 15/30 & 8674 
& 31/40 & 4845 \\
\bottomrule
\end{tabular}
}
\caption{Impact of applying a 12k token cutoff on measured token usage across benchmarks. 
Accuracy remains largely unaffected, while the average token count drops significantly—particularly on harder tasks like AIME and AMC. 
This demonstrates the necessity of outlier mitigation for fair efficiency comparison.}
\label{tab:cutoff_effect}
\end{table*}

As shown in Table~\ref{tab:cutoff_effect}, applying the 12k token cutoff has a significant effect, particularly on harder benchmarks like AIME24 and AMC. On these tasks, the model occasionally fails to generate valid answers and enters degenerate loops, repeatedly producing the same text until reaching the maximum token limit. As a result, the average token usage can increase by up to 5×, which does not reflect the true distribution of reasoning length and severely skews efficiency comparisons.

\section{Fine-Tuning Details}
\label{app:sft}

We fine-tune the DeepSeek-R1-Distill-Qwen models (7B and 1.5B) using the \textsc{LLaMA-Factory} framework with LoRA adaptation. The training data includes 8,000 samples drawn from GSM8K and PRM12K. We set a cutoff length of 4096 tokens for both input and output sequences.

The models are fine-tuned for 3 epochs using the following configuration:

\begin{itemize}
    \item \textbf{Cutoff length:} 4096
    \item \textbf{Max samples:} 8000
    \item \textbf{Batch size:} 2 (with gradient accumulation of 4)
    \item \textbf{Learning rate:} 3e-5 with cosine schedule
    \item \textbf{Precision:} bf16
    \item \textbf{Validation split:} 5\% of training data
    \item \textbf{Evaluation strategy:} every 300 steps
\end{itemize}

Training is performed using two A100 80GB GPU. All experiments use \texttt{overwrite\_cache=true} and 8 parallel preprocessing workers. The resulting models are directly used for downstream evaluation without additional tuning.



\section{Prompt Templates}
\label{appendix:prompts}
The full prompt for skill-based step segmentation:
\begin{tcolorbox}[colback=gray!5, colframe=gray!50, title=Skill-Based Step Segmentation Prompt (Full Version)]
{\slshape
You are given a complete reasoning path for a math problem. Your task is to segment it into a sequence of clear, non-overlapping steps, where each step corresponds to exactly one atomic mathematical skill. These skills may include, for example, addition, subtraction, applying a formula, interpreting a quantity, simplifying, checking a condition, and so on.

Use the following format for each step:

\texttt{Step n: \{\{original text segment\}\}\\Skill: \{\{name of the skill used\}\}}

Only segment and label the steps — do not solve or modify the original content in any way.
}
\end{tcolorbox}

The full prompt for self-revision:

\vspace{1em}

\begin{tcolorbox}[colback=gray!5, colframe=gray!50, title=Self-Revision Prompt (Full Version)]
{\slshape
You are an expert in mathematical reasoning compression.

Given a list of reasoning steps labeled with their respective skills, your task is to evaluate and revise each step according to one of the following four actions:

1. \textbf{KEEP}: The step is necessary and already concise. Keep it unchanged.\\
2. \textbf{DELETE}: The step is unnecessary and should be removed entirely.\\
3. \textbf{SINGLE-STEP COMPRESS}: The step is necessary but verbose; rewrite it in a more concise way.\\
4. \textbf{MULTI-STEP COMPRESS}: The step can be merged with neighboring steps; write a combined, cleaner version.

If the final step clarifies the final answer (e.g., “The answer is...”), retain it.

After completing the revision of each step, synthesize the revised steps into a coherent explanation. Ensure the output is fluent, logically sound, and matches the speaker’s tone and style.
}
\end{tcolorbox}

\section{Skill Base Decomposition Evaluation}
\label{skill eval}

To assess the quality of our skill-based step decomposition, we conduct a pairwise comparison experiment using \textbf{Gemini 2.0 Flash} as the judge~\cite{gao2023human,li2024generation,li2025preference}. For each reasoning trace \( T \), we generate two segmentations:
(1) our proposed \textit{skill-based decomposition}, and
(2) a baseline \textit{sentence-based split} (obtained by naïvely splitting at punctuation or sentence boundaries).

We prompt Gemini to select the version with more \textbf{semantically coherent}, \textbf{structurally stable}, and \textbf{granularity-appropriate} steps, using the following instruction:

\vspace{0.5em}
\begin{quote}
Given two versions of step-by-step reasoning derived from the same original trace, please choose the version that shows better semantic coherence, structural consistency, and appropriate granularity. Respond with `A', `B', or `Same'.
\end{quote}
\vspace{0.5em}

Among 50 randomly sampled traces from GSM8K, Gemini selected our skill-based decomposition in \textbf{33} cases, while preferring the baseline in only \textbf{17} cases. This result suggests that our segmentation method provides more interpretable and structurally aligned reasoning steps, which are beneficial for downstream pruning and distillation.

\section{Skill-based Decomposition vs. Default}
\label{skill_default}

In this section, we show a non-cherry picking example of how our skill based decomposition different from the default one. As shown in Table~\ref{tab:decomposition_comparison}, our method has a more fine-grained step decomposition effect which leads to a more detailed pruning effect.

\begin{table*}[t]
\small
\centering
\begin{tabularx}{\textwidth}{>{\bfseries}p{4.2cm} X}
\toprule
\textbf{Form} & \textbf{Decomposition Output} \\
\midrule

Problem & Joy can read 8 pages of a book in 20 minutes. How many hours will it take her to read 120 pages? \\
\midrule

Original Long-form CoT & 
First, I need to determine Joy’s reading rate. She reads 8 pages in 20 minutes, so her reading rate is 8 pages per 20 minutes.

Next, I’ll calculate how many pages she reads per minute by dividing 8 pages by 20 minutes, which gives 0.4 pages per minute.

Then, to find out how many minutes it takes her to read 120 pages, I’ll divide 120 pages by her reading rate of 0.4 pages per minute, resulting in 300 minutes.

Finally, I’ll convert 300 minutes into hours by dividing by 60, which equals 5 hours. \\
\midrule

DRP Decomposition (Skill-Based) & 
\textbf{Step 1:} Determine Joy’s reading rate. She reads 8 pages in 20 minutes, so her reading rate is 8 pages per 20 minutes. \newline
\textit{Skill: Interpreting a quantity} \newline
\textbf{Step 2:} Calculate how many pages she reads per minute by dividing 8 pages by 20 minutes. \newline
\textit{Skill: Division} \newline
\textbf{Step 3:} The result of the division is 0.4 pages per minute. \newline
\textit{Skill: Simplifying} \newline
\textbf{Step 4:} To find out how many minutes it takes her to read 120 pages, divide 120 pages by 0.4 pages per minute. \newline
\textit{Skill: Division} \newline
\textbf{Step 5:} The result of the division is 300 minutes. \newline
\textit{Skill: Simplifying} \newline
\textbf{Step 6:} Convert 300 minutes into hours by dividing by 60. \newline
\textit{Skill: Division} \newline
\textbf{Step 7:} The result of the division is 5 hours. \newline
\textit{Skill: Simplifying} \\
\midrule

DRP Pruning and Final Result &
1. \textbf{KEEP:} The step is necessary and already concise. \newline
2. \textbf{SINGLE-STEP COMPRESS:} Joy reads 8 pages in 20 minutes, so she reads 0.4 pages per minute (8 ÷ 20). \newline
3. \textbf{DELETE:} This step is now included in the previous step. \newline
4. \textbf{SINGLE-STEP COMPRESS:} To find the time for 120 pages, divide 120 by 0.4 pages per minute. \newline
5. \textbf{DELETE:} This step is now included in the previous step. \newline
6. \textbf{SINGLE-STEP COMPRESS:} Convert 300 minutes to hours by dividing by 60. \newline
7. \textbf{DELETE:} This step is now included in the previous step. \newline

\textbf{Final Answer:} Okay, let's calculate the time it takes Joy to read 120 pages. Joy reads 8 pages in 20 minutes, which means she reads 0.4 pages per minute (8 ÷ 20). To find out how many minutes it takes her to read 120 pages, divide 120 by 0.4, resulting in 300 minutes. Convert 300 minutes into hours by dividing by 60, which gives us 5 hours. The answer is 5 hours. \\
\midrule

Default Decomposition &
\textbf{Step 1:} First, I need to determine Joy's reading rate. She reads 8 pages in 20 minutes, so her reading rate is 8 pages per 20 minutes. \newline
\textbf{Step 2:} Next, I'll calculate how many pages she reads per minute by dividing 8 pages by 20 minutes, which gives 0.4 pages per minute. \newline
\textbf{Step 3:} Then, to find out how many minutes it takes her to read 120 pages, I'll divide 120 pages by her reading rate of 0.4 pages per minute, resulting in 300 minutes. \newline
\textbf{Step 4:} Finally, I'll convert 300 minutes into hours by dividing by 60, which equals 5 hours. \\
\midrule

Default Pruning and Final Result &
\textbf{Step 1:}  SINGLE-STEP COMPRESS \newline
Revised: Joy's reading rate is 8 pages per 20 minutes. \

\textbf{Step 2:} SINGLE-STEP COMPRESS \newline
Revised: Dividing 8 by 20, Joy reads 0.4 pages per minute. 

\textbf{Step 3:} SINGLE-STEP COMPRESS \newline
Revised: Dividing 120 by 0.4, it takes her 300 minutes to read 120 pages. 

\textbf{Step 4:} SINGLE-STEP COMPRESS \newline
Revised: Converting 300 minutes to hours: 300 ÷ 60 = 5 hours. 

\textbf{Final Answer:} \newline
Joy's reading rate is 8 pages per 20 minutes. Dividing 8 by 20, Joy reads 0.4 pages per minute. Dividing 120 by 0.4, it takes her 300 minutes to read 120 pages. Converting 300 minutes to hours: 300 ÷ 60 = 5 hours. \textbf{The answer is 5.} \\
\bottomrule
\end{tabularx}
\caption{A comparison between DRP’s skill-based decomposition and default segmentation with compression, for the same math problem.}
\label{tab:decomposition_comparison}
\end{table*}

\section{Long CoT vs. Short CoT}
\label{long_cot}

In this section, we show a non-cherry picking example of how our skill based decomposition different from the default one. As shown in Table~\ref{tab:cot_trajectory_comparison}, our DRP-style CoT are more informative. That might because of LLM's unfair ability on different aspects~\cite{jiang2024memorization}.

\begin{table*}[ht]
\centering
\renewcommand{\arraystretch}{1.25}
\begin{tabularx}{\textwidth}{l X}
\toprule
\textbf{Form} & \textbf{Reasoning Trajectory} \\
\midrule
\multicolumn{2}{l}{\textbf{Problem:} Joy can read 8 pages of a book in 20 minutes. How many hours will it take her to read 120 pages?} \\
\midrule
\textbf{Original Long-form CoT} & 
First, I need to determine Joy’s reading rate. She reads 8 pages in 20 minutes, so her reading rate is 8 pages per 20 minutes.

Next, I’ll calculate how many pages she reads per minute by dividing 8 pages by 20 minutes, which gives 0.4 pages per minute.

Then, to find out how many minutes it takes her to read 120 pages, I’ll divide 120 pages by her reading rate of 0.4 pages per minute, resulting in 300 minutes.

Finally, I’ll convert 300 minutes into hours by dividing by 60, which equals 5 hours. \\
\midrule
\textbf{DRP-style (Ours)} & 
Okay, let's calculate the time it takes Joy to read 120 pages. Joy reads 8 pages in 20 minutes, which means she reads 0.4 pages per minute ($8 \div 20$). To find out how many minutes it takes her to read 120 pages, divide 120 by 0.4, resulting in 300 minutes. Convert 300 minutes into hours by dividing by 60, which gives us 5 hours. The answer is 5 hours. \\
\midrule
\textbf{Short-form CoT} & 
In one hour, there are 3 sets of 20 minutes. So, Joy can read $8 \times 3 = 24$ pages in an hour. It will take her $120 \div 24 = 5$ hours to read 120 pages. \\
\bottomrule
\end{tabularx}
\caption{A comparison between original long-form, DRP-style, and short-form reasoning traces for the same math problem.}
\label{tab:cot_trajectory_comparison}
\end{table*}

\end{document}